\pdfoutput=1
\documentclass[11pt]{article}

\usepackage[]{acl}

\usepackage{times}
\usepackage{latexsym}
\usepackage{amsmath}
\usepackage{hyperref}

\usepackage[T1]{fontenc}
\usepackage[utf8]{inputenc}
\usepackage{graphicx}
\usepackage{microtype}
\usepackage{graphicx}

\usepackage{multirow}
\usepackage{algorithm}
\usepackage{algorithmicx}
\usepackage{algpseudocode} 
 
\usepackage{algpseudocode}
\usepackage{bbding}
\usepackage{booktabs}
\usepackage{longtable}
\title{KLoB: a Benchmark for Assessing Knowledge Locating Methods \\ in Language Models}

\author{
    Yiming Ju\textsuperscript{1}, \ \
    Huimin Ma\textsuperscript{}, \ \
    Xingrun Xing\textsuperscript{1},  \ \ 
    \textbf{Zhixiong Zeng\textsuperscript{2}} 
    \\
    \textsuperscript{\rm 1} Beijing Academy of Artificial Intelligence\\
    \textsuperscript{\rm 2} Tencent \\
    \emph{yiming.ju2024@gmail.com,  \ \  m163mhm@163.com} \\ \emph{xingxingrun2023@ia.ac.cn, \ \ barretzeng@tencent.com} \\
}

\begin{document}
\maketitle

\begin{abstract}
Recently, Locate-Then-Edit paradigm has emerged as one of the main approaches in changing  factual knowledge stored in the Language models.
However, there is a lack of research on whether present locating methods can pinpoint the exact parameters embedding the desired knowledge.
Moreover, although many researchers have  questioned the validity of  locality hypothesis of factual knowledge, no method is provided to test the a hypothesis for more in-depth discussion and research.
Therefore, we introduce KLoB, a benchmark examining three essential properties that a reliable knowledge locating method should satisfy.
KLoB can serve as a benchmark for evaluating existing locating methods in language models, and can
contributes a method to reassessing the validity of locality hypothesis of factual knowledge.
KLoB is publicly available at an anonymous GitHub: \url{https://github.com/anon6662/KLoB}.
% KLoB is publicly available at \url{https://github.com/juyiming/KLoB}.

\end{abstract}

\section{Introduction}

% Language models have exhibited a significant capability to store factual knowledge \citep{roberts2020much}. Yet, as language models scale larger, the need to uphold the correctness and contemporaneity of factual knowledge
% % , without incurring excessive retraining costs, 
% becomes increasingly critical \citep{sinitsin2020editable}， 
% % This burgeoning need paves the way for a novel area of research, namely knowledge editing.
% and spark  a new area of research：knowledge editing.
Language models have exhibited a significant capability to store factual knowledge \citep{roberts2020much}. Yet, as language models scale larger, the need to uphold the correctness and contemporaneity of stored knowledge becomes increasingly critical \citep{sinitsin2020editable}, thus sparking a new area of research: knowledge editing.
Among current knowledge editing techniques, \textbf{Locate-Then-Edit} paradigm \citep{dai2022knowledge, meng2022locating, meng2022mass,li2023pmet} has emerged as one of the main approaches and garnered significant attention \citep{yao2023editing}. 

\begin{figure}[t!]
\centering
\includegraphics[width=0.99\linewidth]{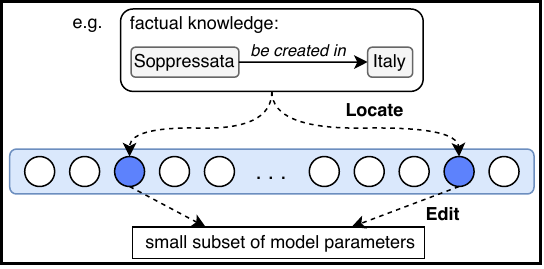}
\caption{Illustration of Locate-Then-Edit method.
}
\label{locality hypothesis}	
\end{figure}

As depicted in Figure \ref{locality hypothesis}, by first locating parameters associated with specific knowledge and modifying while keeping the remaining parameters unchanged,  Locate-Then-Edit methods can facilitate alterations to the model with very low cost \citep{yao2023editing}.
However, there is currently no method for evaluating the locating results.
It's still ambiguous whether current locating methods can pinpoint the exact parameters embedding the desired knowledge. 
Moreover, the \textbf{locality hypothesis of factual knowledge}, which posits that factual knowledge is predominantly embedded within a small subset of parameters, 
has encountered a degree of skepticism and warrants further investigation.
Yet, there's a noticeable absence of established methods to study and validate this concern.

 %has also faced skepticism and remains to be investigated.
% Yet, there's a noticeable absence of established methods to address and validate these concerns.

% The Locate-Then-Edit methods are based on locality hypothesis of factual knowledge.

% As depicted in Figure \ref{locality hypothesis}, this hypothesis posits that factual knowledge is predominantly embedded within a small subset of the model's parameters.

% This understanding suggests that model knowledge can be modified by changing a limited set of parameters. 

% However, the validity of the assertion that factual knowledge primarily resides within a small subset of the model's parameters — the core of the locality hypothesis of factual knowledge — remains to be rigorously investigated.

% Moreover, even assuming the locality hypothesis is valid, it's still ambiguous whether present locating methods can pinpoint the exact parameters embedding the desired knowledge. 

% Furthermore, assessing how well the identified parameters can encapsulate the target knowledge in comparison with other parameters, is an important and intriguing avenue for study. Yet, there's a noticeable absence of established methods to address and validate these concerns.

\begin{figure*}[t!]
\centering
\includegraphics[width=0.99\linewidth]{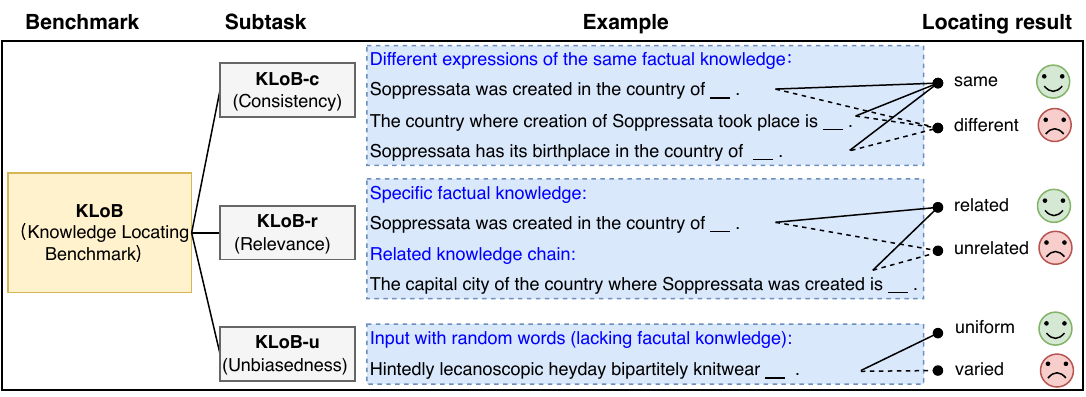}
\caption{Examples of three subtasks in KLoB. 
KLoB-c: Each example comprises multiple expressions of the same factual knowledge;
KLoB-r: Each example includes a sentence with specific factual knowledge and another with a related knowledge chain;
KLoB-u: Each example features a sentence composed of random words.
}
\label{KLOB}	
\end{figure*}

Therefore, we introduce KLoB (\textbf{K}nowledge \textbf{Lo}cating \textbf{B}enchmark), a novel benchmark for evaluating locating methods in language models. 
KLoB delineates three essential criteria that a reliable knowledge locating method should satisfy and then evaluates the efficacy of locating methods by examining these criteria. The delineated criteria are as follows:

\begin{itemize}
    \item \textbf{Consistency}: Locating results should remain consistent across different expressions of the same factual knowledge.
    \item \textbf{Relevance}: Locating results for related factual knowledge should exhibit higher similarity than those for unrelated knowledge.
    \item \textbf{Unbiasedness}: Parameter scores should be more uniform for inputs lacking explicit factual knowledge than those for inputs with explicit knowledge.
\end{itemize}

As illustrated in Figure \ref{KLOB}, the KLoB benchmark comprises three subtasks, each examining one of the aforementioned criteria.
As the first benchmark for evaluating knowledge locating methods in language models, KLoB can play a crucial role in facilitating a comprehensive evaluation of whether current locating methods can accurately pinpoint model parameters associated with specific factual knowledge.
Furthermore, KLoB can also contribute a method to study and reassess the validity of the locality hypothesis of factual knowledge.

\section{Preliminary}

\subsection{Knowledge Editing}

The factual knowledge embedded within a language model can be incorrect or become outdated over time.
Knowledge editing aims to hold the correctness and contemporaneity of factual knowledge, without incurring excessive retraining costs.
Current knowledge editing methods can be mainly categorized into three ways \citep{yao2023editing}:
\begin{itemize}
    \item Locate-Then-Edit: Locating and modifying model parameters that are associated with specific knowledge \citep{dai2022knowledge, meng2022locating, meng2022mass, li2023pmet}.
    \item Meta-learning: Utilizing a hyper network to learn how to change parameters of a language model \citep{de2021editing,mitchell2021fast}.
    \item Memory Model: Storing edits in an additional model while keeping the original model frozen \citep{mitchell2022memory,huang2023transformer,dong2022calibrating}.
\end{itemize}

\subsection{Locate-Then-Edit Method}
Based on the locality hypothesis of factual knowledge, which posits that knowledge is primarily embedded within a subset of the model's parameters, 
Locate-Then-Edit methods operate in a pipeline manner: first locate a small subset of model parameters, which are associated with specific knowledge, and then modify those parameters. 
Because Locate-Then-Edit methods only change the located parameters while keeping the rest unchanged, they can effectively modify the model in a more targeted manner and have garnered significant attention.

\section{KLoB: Knowledge Locating Benchmark}

\subsection{Design Philosophy}
% KLoB is designed for evaluating whether parameters selected by locating methods embed the desired knowledge.
% The philosophy of construction of KLoB is examining whether the locating results meet the desired properties.
% Assuming the validity of the locality hypothesis of factual knowledge, we
% define three criteria that a reliable knowledge locating
% method should satisfy: Consistency, Relevance, and Unbiasedness.

KLoB is designed for evaluating whether parameters selected by locating methods embed the desired knowledge. 
The underlying philosophy in constructing KLoB is to examine whether the locating results possess the desired properties. Assuming the validity of the locality hypothesis of factual knowledge, we define three criteria that a reliable knowledge locating method should satisfy: Consistency, Relevance, and Unbiasedness.

\subsubsection{Consistency}
Since the goal of locating methods is selecting parameters associated with specific factual knowledge, the locating result should be associated solely with the targeted knowledge, and should not be affected by other factors such as syntactic structure or synonym substitution. 
Therefore, we propose consistency as a criterion: \textbf{for the same factual knowledge, locating results should remain invariant despite variations in expression}.

% Therefore, we propose consistency as a property that knowledge locating results should adhere to: \textbf{for the same factual knowledge, it should remain invariant despite variations in expression}.

% Therefore, we propose relevance as a criterion: \textbf{the locating results for specific knowledge and a related knowledge chain that includes it should exhibit greater similarity compared to those for unrelated knowledge}.

\subsubsection{Relevance}
\citet{huang2023emotionally} introduced the concept of multi-hop knowledge editing, where an input may be linked to a chain of interconnected factual knowledge. 
Locating methods should be able to recognize the correlation between the specific knowledge and the knowledge chain that includes it.
Therefore, we propose relevance as a criterion: \textbf{the locating results for specific knowledge and its related knowledge chain should exhibit greater similarity compared to those for unrelated knowledge}.

\subsubsection{Unbiasedness}
% Knowledge locating methods score and rank parameters according to the factual knowledge. 
Knowledge locating methods score and rank parameters based on their association with the targeted knowledge.
Since the differences in parameter scores arise from the knowledge present in the input,
for inputs that do not align with any factual knowledge, the parameter scores should be more uniform
compared to those aligned with specific factual knowledge.
Therefore, we propose unbiasedness as a criterion: \textbf{compared to inputs explicitly pointing to factual knowledge, parameter scores for inputs devoid of factual knowledge should be more uniform}.

\subsection{Data Format}

As depicted in Figure \ref{KLOB}, KLoB consists of three subtasks,  each examining one of the aforementioned criteria:

\begin{itemize}
    % \item KLoB-c (consistency): each example comprises multiple expressions of the same factual knowledge. 
    % As shown in Figure \ref{KLOB}, \emph{'Soppressata was created in the country of _'} and \emph{'The country where creation of Soppressata took place is '} both ponits to the factual kownledge \emph{(Soppressata $\xrightarrow{\text{created in}}$ Italy)}.
    % The evaluation metrics of KLoB-c is whether the locating results of these expressions are same. 

        \item KLoB-c (consistency): In this subtask, each example comprises three sentences of the same factual knowledge. As shown in Figure \ref{KLOB}, both \emph{'Soppressata was created in the country of \_ '} and \emph{'The country where the creation of Soppressata took place is \_'} include the factual knowledge [Soppressata $\xrightarrow{\text{created in}}$ Italy].
        
        \item KLoB-r (relevance): Here, each example comprises a sentence that includes specific factual knowledge and another one associated with its related knowledge chain. As depicted in Figure \ref{KLOB}, the sentences  
        \emph{'Soppressata was created in the country of \_'} 
         and \emph{'The capital city of the country where Soppressata was created is \_'} correspond to the factual knowledge [Soppressata $\xrightarrow{\text{created in}}$ Italy] and its related knowledge chain [Soppressata, $\xrightarrow{\text{created in}}$ Italy, $\xrightarrow{\text{capital}}$ Rome)] respectively.

        \item KLoB-u (unbiasedness): Each example in this subtask features a sentence composed of random words, which is considered devoid of factual knowledge.
\end{itemize}

As depicted in Figure \ref{KLOB}, the answer entity in factual knowledge is positioned at the end of the sentence in KLoB. This design distinguishes KLoB from previous benchmarks \citep{elazar2021measuring, huang2023emotionally} that utilize sentences with ['MASK'] or question sentences as input text. Consequently, examples in KLoB are compatible with both auto-regressive models, such as GPT \citep{radford2018improving, radford2019language} and Llama \citep{touvron2023llama}, and autoencoding models, such as BERT \citep{kenton2019bert} and ALBERT \citep{lan2019albert}.

\begin{table}[t!]
\centering
\begin{tabular}{lccc}
\toprule[0.8pt]
subtask & relations  & avg length & examples \\
\midrule[0.4pt]
KLoB-c & 32  & 8.75 & 13675 \\
% \midrule[0.4pt]
KLoB-r & 35  & 20.4 & 9548 \\
% \midrule[0.4pt]
KLoB-u & /  & 10.1 & 25470 \\
\bottomrule[0.8pt]
\end{tabular}
\caption{Data statistics of KLoB benchmark.}
\label{ana}
\end{table}

\subsection{Data Construction}
KLoB is constructed based on Wikidata\citep{vrandevcic2014wikidata} and  MQUAKE benchmark \cite{zhong2023mquake}. Table \ref{ana} summarizes the statistics of KLoB benchmark.

% %, and manually crafted templates.
% PARAREL \citep{elazar2021measuring}
%\subsubsection{Construction of KLoB-c}
\begin{itemize}
    % \item KLoB-c is constructed based on Wikidata, a knowledge base consisting of millions of fact triples. 
    % We select relationships and corresponding facts in Wikidata, and mannully construct 3 templates for each relationship.
    % The template was maually wrriten by human experts , ensuring diversity in grammatical structures and clear pointing to a specific entity.\stepcounter{footnote}\footnotetext{Some relationship templates may lead to ambiguity with certain facts. For example, consider the template '\textbf{[X]}  is held in \textbf{[Y]} '; when \textbf{[X]}  is 'the 28th Summer Olympic Games', \textbf{[Y]}  might be 'Greece' or 'Athens'. To avoid this kind of ambiguity, the entity type is specified within some template. For example, the above template is be converted to '\textbf{[X]}  is held in the country of \textbf{[Y]} '.}\footnotemark[1] Table \ref{templates} in the Appendix shows the maually created templates for constructing KLoB-c.
    
    % \item KLoB-c is built upon Wikidata, a knowledge base consisting of millions of factual triples. We select relationships from Wikidata and manually construct three templates for each relationship. These templates are manually written by human experts, ensuring diversity in grammatical structures and words.
    % Table \ref{templates} in the Appendix shows all manually created templates for constructing KLoB-c. Then we use these tempaltes and corresponding entities in Wikidata to generate examples.

    \item \textbf{KLoB-c} is built upon Wikidata, a knowledge base consisting of millions of factual triples. We select relationships from Wikidata and manually construct three templates for each relationship. These templates are manually written by human expert, ensuring diversity in grammatical structures and words. Table \ref{templates} in the Appendix lists all templates for constructing KLoB-c and  Table \ref{relation-counts} shows example counts for each relation.\footnote{There are 96 templates and 32 relations, with each relation corresponding to 3 templates.} Then, we use these templates and corresponding entities in Wikidata to generate sentences. 

\begin{table*}[htb]
\scriptsize
\centering
\resizebox{0.9\linewidth}{!}{
\begin{tabular}{l c | l c}
\toprule[0.8pt]
\textbf{Relation} & \textbf{Example Counts} & \textbf{Relation} & \textbf{Example Counts} \\
\midrule[0.4pt]
\textbf{P103} (native) & 851 & \textbf{P276} (locate) & 324 \\
\textbf{P1001} (legal-term) & 526 & \textbf{P36} (capital) & 377 \\
\textbf{P101} (work) & 198 & \textbf{P37} (official-language) & 667 \\
\textbf{P106} (by-profession) & 292 & \textbf{P39} (position) & 280 \\
\textbf{P108} (works-for) & 263 & \textbf{P407} (write) & 662 \\
\textbf{P127} (owned-uy) & 330 & \textbf{P413} (play) & 649 \\
\textbf{P1303} (play) & 264 & \textbf{P449} (air) & 567 \\
\textbf{P131} (located-in) & 137 & \textbf{P495} (create) & 649 \\
\textbf{P136} (plays-music) & 521 & \textbf{P26} (spouse) & 266 \\
\textbf{P1376} (capital) & 140 & \textbf{P50} (author) & 828 \\
\textbf{P138} (is-name-after) & 214 & \textbf{P112} (founded by) & 181 \\
\textbf{P1412} (communicate) & 120 & \textbf{P69} (educated at) & 116 \\
\textbf{P159} (headquarter) & 590 & \textbf{P140} (affiliated-with) & 349 \\
\textbf{P17} (is-located) & 783 & \textbf{P175} (performer) & 386 \\
\textbf{P176} (produce) & 831 & \textbf{P641} (sport) & 367 \\
\textbf{P178} (develop) & 702 & \textbf{P19} (born) & 245 \\
\bottomrule[0.8pt]
\end{tabular}
}
\caption{Number of examples of each relation in KLoB-c}
\label{relation-counts}
\end{table*}

    \item \textbf{KLoB-r} is constructed based on MQUAKE, a dataset comprising multi-hop questions corresponding to chains of facts. We select two-hop fact chains from MQUAKE and use the first fact in chains to generate sentences containing single fact, employing the same template as utilized in KLoB-c. For entire fact chains, sentences are generated by rephrasing the multi-hop questions in MQUAKE.
\\
    \item \textbf{KLoB-u} is constructed by replacing the words in the examples from KLoB-c and KLoB-r with random words and punctuation. The replacement words are sourced from the English word list of the NLTK (Natural Language Toolkit) library \citep{bird2009natural}.

\end{itemize}

Since our goal is to study the knowledge embedded in language models, we filter out factual knowledge that is difficult for language models to recall.
Specifically, we query Llama2-7b \citep{touvron2023llama}  using an in-context learning approach with 8 demonstration examples and retain only the factual knowledge for which the model can correctly predict the answers.

% \vspace*{\fill}  % 这使得脚注文本出现在页面底部
% \footnotetext{Detailed formula for calculating RSim can be found in the Appendix.}
% \stepcounter{footnote}
% \footnotetext{}

\begin{figure}[t!]
\centering
\includegraphics[width=0.95\linewidth]{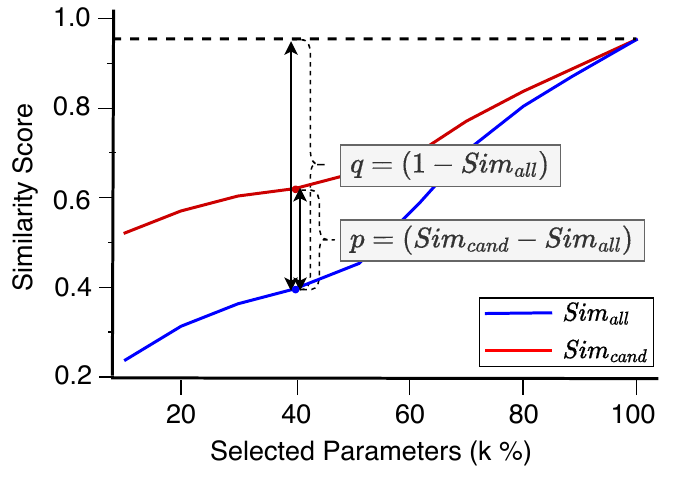}
\caption{$RSim$: evaluation metrics for KLoB-c and KLob-r. 
$RSim$ is given by $RSim = max(\frac{p}{q}, \ 0)$. 
}
\label{eval}	
\end{figure}
% by GPT-J\footnote{We select only the factual knowledge that GPT-J could recall.}.
    
\subsection{Evaluation Metrics
}
% We employ the following evaluation metrics to quantify the performance of the knowledge locating method across three subtasks in KLoB. 
% The evaluation for KLoB-c and  KLoB-r is the similarity between locating results. 

Due to the varying granularities of localization used in different knowledge editing approaches (ranging from a single neuron \citep{dai2022knowledge} to a complete model layer \citep{meng2022locating}), and considering the different extents of parameters involved, some methods might analyze the entire model's parameters \citep{dai2022knowledge, meng2022locating, meng2022mass} while others focus only on a subset \citep{li2023pmet}. Consequently, calculating the similarity solely among chosen parameters does not provide a fair basis for comparing the multitude of knowledge localization methods across a uniform metric. 
To address this issue, we have designed relative metrics to serve as an evaluation standard for the KLoB benchmark: Relative Similarity (RSim) for KLoB-c and KLoB-r, and Relative Standard Deviation (RSD) for KLoB-u.

% similarity between two locating results by comparing the

% similarity of the locating result under evaluation with that of the entire dataset.

% The evaluation criteria for both KLoB-c and KLoB-r involve assessing the similarity between the locating results.

% are whether the locating results of sentence in a example are the same.
%   The evaluation metrics for KLoB-r are whether the locating results of these sentences are related.
% The evaluation metric is whether the scores of model parameters are uniform.

% In this work, we design two evaluation metrics for subtasks in KLoB. 

\begin{itemize}
    \item \textbf{Relative Similarity}: RSim metric assesses the extent to which the similarity within certain localization results exceeds the similarity between these and other localization results, operating by considering only the selected parameters in localization results.
    The first step in the RSim metric is to calculate the intra-similarity of the candidate localization results (e.g., localization results for sentences in one KLoB-c example), which is denoted as $Sim_{cand}$. Then, RSim calculates the similarity between the candidate localization results and localization results of all samples in the subtask, which is denoted as $Sim_{all}$.   
     As depicted in Fig. \ref{eval}, the RSim metric is is calculated using the formula: $RSim = max(\frac{Sim_{cand} - Sim_{all}}{1 \ - \ Sim_{all}}, \ 0)$. 
     Given that the upper bound of similarity value $Sim$ is 1, this formula quantifies the extent to which $Sim_{cand}$ is closer to the upper bound compared to $Sim_{all}$. 
     If the candidate localization results are all identical,  meaning that $Sim_{cand}$ equals 1, then RSim is equal to 1. 
     If $Sim_{cand}$ is less than or equal to $Sim_{all}$, then RSim is equal to 0. 
     The detailed formula for calculating RSim is elaborated in the Appendix, Section A.
     
     % \footnotemark
     % \footnote{Detailed formula for calculating RSim can be found in the Appendix.}
    \item \textbf{Relative Standard Deviation}: 
    RSD metric assesses how much more uniform the localization results are for sentences devoid of factual knowledge (KLoB-u) compared to those with explicit factual knowledge (KLoB-c and KLoB-r), operating by considering all model parameters. RSD utilizes the standard deviation to quantify the variability of parameter scores in one localization result, which first calculates the average SD of sentences in KLoB-c and KLoB-r (sentences with factual knowledge), denoted as SD$_{factual}$. Then, it calculates the average SD of sentences in KLoB-u (sentences without factual knowledge), denoted as SD$_{nonfactual}$. RSD is given by $RSD = \max\left(1 - \frac{SD_{nonfactual}}{SD_{factual}}, \ 0\right)$. 
RSD measures the relationship between of SD$_{nonfactual}$ and  SD$_{factual}$ by calculating the proportion between them.
If the parameter scores for sentences without knowledge are all identical, SD$_{nonfactual}$ equals 0, which means RSD equals 1. 
Conversely, if the parameter scores for sentences without knowledge are more variable than those with factual knowledge, meaning that $SD_{nonfactual} > SD_{factual}$, then RSD equals 0.

    % This formula normalizes the standard deviation of the scores using a ratio relationship. 
    % If the parameter scores for sentence without knowledge are identical, $SD_{nk}$ equals 0, which means RSD equals 1. 
    % If the parameter scores for sentence without knowledge are more distinguishable compared to sentnce with factual knowledge, $SD_{nk}$ is greater than $SD_{k}$, then RSD equals 0.

    % \footnotemark[2]
\end{itemize}

% \subsection{Data Analysis}

% \section{Experiment}
% This section introduces the experiments conducted on KLoB.
% \subsection{Locating Methods and Models
% }
% We conduct evaluations on KLoB
% utilizing two prominent knowledge locating methodologies:

% \begin{itemize}
%     \item \textbf{Knowledge Attribution} \citep{} evaluates the contribution of each neuron to knowledge predictions based on integrated gradients \citep{sundararajan2017axiomatic}.
%     \item \textbf{Causal Tracing} \citep{} evaluates the contribution of a unit towards a factual prediction through results from three runs: clean run, corrupted run and corrupted-with-restoration run.
%     % refers to \emph{ causal mediation analysis} \citep{}, which 
%     % calculate each state’s contribution towards a correct factual prediction.
% \end{itemize}

% The experiments are conducted based on four models:
% BERT-uase, BERT-large \citep{}, GPT-2 \citep{} and GPT-neo \citep{}.

% \subsection{Experimental Results}

\section{Conlusion}
In this work, we introduce KLoB, the first benchmark for evaluating locating methods in language models.
KLoB delineates three essential properties a reliable knowledge locating method should satisfy: Consistency, Relevance, and Unbiasedness. Thus, the evaluation of locating methods can be conducted by examining these properties.
We hope KLoB can serve as a benchmark for evaluating existing locating methods, and contributes a quantitative analysis methods for reassessing the validity of locality hypothesis of factual knowledge.

% \section{Limitations}
% There are two primary limitations in this study that remain unexplored:
% \textbf{(i)} KLoB is only applicable for comparing knowledge locating methods with the same parameter granularity. 
% For instance,  a method with neuron-level granularity versus another at a model layer level cannot be effectively compared using KLoB. 
% \textbf{(ii)} While this work introduces KLoB for evaluating the effectiveness of knowledge locating methods,
% it did not conduct experiments based on existing methods, %which will form the crux of our subsequent research efforts.
% which will be the primary focus of our subsequent research efforts.

% There are two primary limitations in this study that remain unexplored:
% (i) KLoB's Applicability: KLoB is tailored for comparing knowledge locating methods only at identical levels of parameter granularity. For instance, a method with neuron-level granularity versus another at a model layer level cannot be effectively compared using KLoB.
% (ii) Lack of Experimental Validation: While KLoB has been introduced as a tool for evaluating knowledge locating methods, this work has not yet conducted experimental validation using existing methods. Addressing this gap will form the crux of our subsequent research efforts."

\bibliography{anthology,custom}

\appendix

% \section{Appendix}
% \subsection{Experimental Details}
% The implementations of Knowledge Attribution \citep{} and Causal Tracing \citep{} are based on their open-source code\footnote{Knowledge Attribution: // Causal Tracing: }. 
% The implementation of Causal Tracing in our experiment has some differences compared to the original version: while the original version employs the whole neural network layer as the granularity for location, we have chosen to use the output values in each individual layer for consistency with the implementation of Knowledge Attribution. 

% \subsection{Detailed Description of Data Format and Construction}

% ~\\
% \noindent
% \textbf{KLoB-c \ }
% The number of prompts in each KLoB-c instance is three. All prompts are constructed from templates maually wrriten by human experts.
% Different form many temolates in PARAREL dataset, templates in KLoB-c have different syntactic structures, rather than merely simple synonym substitutions.

% ~\\
% \noindent
% \textbf{KLoB-r \ }
% Prompts for each premise knowledge in KLoB-r number three, utilizing the same templates as in KLoB-c.\footnote{Samples with relations not found in the templates of KLoB-c are discarded.} 
% Each sample in KLoB-r contains only one prompt for the knowledge chain. This is due to the fact that prompts for the knowledge chain are rephrased from the questions in MQUAKE,
% and the number that can be successfully rewritten is not fixed. If multiple prompts emerge for a single knowledge chain, they are divided into multiple samples.

\section{Detailed Description for calculating RSim}\label{RSim}
% This section describe the the calculation process of evaluation metrics RSim.

% \begin{itemize}
%     \item \textbf{RSim: } 
    Suppose we have two localization results $x_{i}$ and $x_{j}$, each contains $k\%$ of the  model parameters. The similarity between these two localization results is given by $ Sim = overlap_{ij}/(N*k\%)$, where $overlap_{ij}$ denotes the number of overlapping parameters between $x_{i}$ and $x_{j}$, $N$ is the total number of model parameters.
    For each example in KLoB-c, comprising three sentences, the similarity  $Sim_{cand}$ is calculated  as the average of the pairwise similarities among these sentences."
    Similarly, $Sim_{all}$ represents the average similarity between the selected localization result $x_{cand}$ and all other localization results in the subtask, and is defined  as $Sim_{all} = \frac{1}{M} \sum_{i = 1}^{M} Sim(x_{i}, x_{cand})$, where $M$ denotes the number of examples in the subtask. 
    To streamline the calculations, we approximate $Sim_{all}$ as $Sim(x_{all}, x_{cand})$, where $x_{all}$ refers to a localization result based on the average parameter scores of all examples. Therefore, we only need to calculate the similarity score once to get $Sim_{all}$ for one example.
    
    % To simplify calculations, $Sim_{all} \approx Sim(x_{all}, x_{cand})$, where $x_{all}$ is a locating result, in which parameters are selected based on the average parameter score obtained on the entire subtask.
    % % \item \textbf{RSD: } 

% \end{itemize}

\section{Templates of KLoB-c}
Table \ref{templates} shows the relationship templates in KLoB-c.

% \begin{table}[t!]
% \centering
% \begin{tabular}{l | c}
% \toprule[0.8pt]
% \textbf{relation} & \textbf{example counts} \\
% \midrule[0.4pt]
% \textbf{P103} \ (native) & 851 \\
% \textbf{P1001}  \ (legal-term) & 526 \\
% \textbf{P101}  \ (work) & 198 \\
% \textbf{P106}  \ (by-profession) & 292 \\
% \textbf{P108}  \ (works-for) & 263 \\
% \textbf{P127}  \ (owned-uy) & 330 \\
% \textbf{P1303}  \ (play) & 264 \\
% \textbf{P131}  \ (located-in) & 137 \\
% \textbf{P136}  \ (plays-music) & 521 \\
% \textbf{P1376}  \ (capital) & 140 \\
% \textbf{P138}  \ (is-name-after) & 214 \\
% \textbf{P1412}  \ (communicate) & 120 \\
% \textbf{P159}  \ (headquarter) & 590 \\
% \textbf{P17}  \ (is-located) & 783 \\
% \textbf{P176}  \ (produce) & 831 \\
% \textbf{P178}  \ (develop) & 702 \\
% \textbf{P19}  \ (born) & 245 \\
% \textbf{P276}  \ (locate) & 324 \\
% \textbf{P36}  \ (capital) & 377 \\
% \textbf{P37}  \ (official-language) & 667 \\
% \textbf{P39}  \ (position) & 280 \\
% \textbf{P407}  \ (write) & 662 \\
% \textbf{P413}  \ (play) & 649 \\
% \textbf{P449}  \ (air) & 567 \\
% \textbf{P495}  \ (create) & 649 \\
% \textbf{P26}  \ (spouse) & 266 \\
% \textbf{P50  }  \ (author) & 828 \\
% \textbf{P112  }  \ (founded by) & 181 \\
% \textbf{P69  }  \ (educated at) & 116 \\
% \textbf{P140}   \ (affiliated-with) & 349 \\
% \textbf{P175}  \ (performer) & 386 \\
% \textbf{P641}  \ (sport) & 367 \\
% \bottomrule[0.8pt]
% \end{tabular}
% \caption{Number of examples of each relation in KLoB-c}
% \label{relation-counts}
% \end{table}

\begin{table}[t!]
\centering
\scriptsize
\begin{tabular}{p{0.7cm}|p{5.8cm}}
\toprule[0.8pt]
\textbf{relation} & \textbf{ \ \ \ \ \ \ \ \ \ \  \  \ \ \ \ \ \ \ \ \ \ \ \  \ \ \ \ \ \ \ \ \ templates} \\
\midrule[0.4pt]
\multirow{3}{*}{\textbf{P103} }
&The native language of \textbf{\textbf{[X]} } is \textbf{[Y]}  \\ 
&The mother tongue of \textbf{[X]}  is \textbf{[Y]}  \\ 
&The language \textbf{[X]}  speaks at hometown is \textbf{[Y]}  \\
\midrule[0.4pt]
\multirow{3}{*}{\textbf{P1001}}
&\textbf{[X]}  is a legal term in \textbf{[Y]}  \\ 
&\textbf{[X]}  serves as one of the legal term for \textbf{[Y]}  \\ 
&\textbf{[X]} : the legal term for \textbf{[Y]}  \\
\midrule[0.4pt]
\multirow{3}{*}{\textbf{P101} }
&\textbf{[X]}  works in the field of \textbf{[Y]}  \\ 
&\textbf{[X]}  specializes in the field of \textbf{[Y]}  \\ 
&The domain of activity of \textbf{[X]}  is \textbf{[Y]}  \\
\midrule[0.4pt]
\end{tabular}

 \begin{tabular}{p{0.7cm}|p{5.8cm}}
\multirow{3}{*}{\textbf{P106} }
&\textbf{[X]}  works as \textbf{[Y]}  \\ 
&\textbf{[X]} 's occupation is \textbf{[Y]}  \\ 
&the profession of \textbf{[X]}  is \textbf{[Y]}  \\ 
\midrule[0.4pt]
\multirow{3}{*}{\textbf{P108} }
&\textbf{[X]}  works for the company: \textbf{[Y]}  \\ 
&The company that employs \textbf{[X]}  is \textbf{[Y]}  \\ 
&The company providing employment to \textbf{[X]}  is \textbf{[Y]}  \\ 
\midrule[0.4pt]
\multirow{3}{*}{\textbf{P127} }
&\textbf{[X]}  belongs to \textbf{[Y]}  \\ 
&the company that owns \textbf{[X]}  is \textbf{[Y]}  \\ 
&the owner company of \textbf{[X]}  is \textbf{[Y]}  \\ 
\midrule[0.4pt]
\end{tabular}

 \begin{tabular}{p{0.7cm}|p{5.8cm}}
\multirow{3}{*}{\textbf{P1303}}        
&\textbf{[X]}  plays the musical instrument known as the \textbf{[Y]}  \\ 
&\textbf{[X]}  is known for playing the \textbf{[Y]}  \\ 
&In the hands of \textbf{[X]} , music emerges from the \textbf{[Y]}  \\
\midrule[0.4pt]
\multirow{3}{*}{\textbf{P131}  }       
&\textbf{[X]}  is located in \textbf{[Y]}  \\ 
&\textbf{[X]}  can be found in \textbf{[Y]}  \\ 
&The location of \textbf{[X]}  is \textbf{[Y]}  \\
\midrule[0.4pt]
\multirow{3}{*}{\textbf{P136} }  
&\textbf{[X]}  composes in the genre of \textbf{[Y]}  \\  
&\textbf{[X]}  engages in the performance of \textbf{[Y]}  \\ 
&The music genre that \textbf{[X]}  performs is \textbf{[Y]}  \\ 
\end{tabular}

 \begin{tabular}{p{0.7cm}|p{5.8cm}}
\midrule[0.4pt]
\multirow{3}{*}{\textbf{P1376} }  
&\textbf{[X]}  is the capital of \textbf{[Y]}  \\ 
&\textbf{[X]}  holds the status of being the capital of \textbf{[Y]}  \\ 
&\textbf{[X]} , the capital city of \textbf{[Y]}  \\ 
\midrule[0.4pt]
\multirow{3}{*}{\textbf{P138}  } 
&\textbf{[X]}  is the capital of \textbf{[Y]}  \\ 
&\textbf{[X]}  holds the status of being the capital of \textbf{[Y]}  \\ 
&\textbf{[X]} , the capital city of \textbf{[Y]}  \\ 
\midrule[0.4pt]
\multirow{3}{*}{\textbf{P1412} }  
&\textbf{[X]}  used to communicate in the language of \textbf{[Y]}  \\  
&\textbf{[X]}  expressed himself through the language of \textbf{[Y]}  \\ 
&In language \textbf{[X]}  utilizes for communication is \textbf{[Y]}   \\
\end{tabular}

 \begin{tabular}{p{0.7cm}|p{5.8cm}}
\midrule[0.4pt]
\multirow{3}{*}{\textbf{P159} }  
&The headquarters of \textbf{[X]}  is located in \textbf{[Y]}  \\  
&\textbf{[X]} , whose headquarters is in \textbf{[Y]}  \\ 
&\textbf{[X]}  has established its headquarters in \textbf{[Y]}  \\ 
\midrule[0.4pt]
\multirow{3}{*}{\textbf{P17}  } 
&The headquarters of \textbf{[X]}  is located in \textbf{[Y]}  \\ 
&\textbf{[X]} , whose headquarters is in \textbf{[Y]}  \\ 
&\textbf{[X]}  has established its headquarters in \textbf{[Y]}  \\ 
\midrule[0.4pt]
\multirow{3}{*}{\textbf{P176}   }
&\textbf{[X]}  is produced by the company: \textbf{[Y]}  \\ 
&The company behind \textbf{[X]}  Lumia 800 \textbf{[Y]}  \\ 
&\textbf{[X]}  is one among products crafted by \textbf{[Y]}  \\ 
\end{tabular}
\begin{tabular}{p{0.7cm}|p{5.8cm}}
\midrule[0.4pt]
\multirow{3}{*}{\textbf{P178}  } 
&\textbf{[X]} , a product manufactured by the company: \textbf{[Y]}  \\ 
&The company that developed \textbf{[X]}  is \textbf{[Y]}  \\ 
&The company that  stands behind the creation of \textbf{[X]}  is \textbf{[Y]}  \\ 
\midrule[0.4pt]
\multirow{3}{*}{\textbf{P19}   }
&\textbf{[X]}  was born in \textbf{[Y]}  \\ 
&The place of birth for \textbf{[X]}  is \textbf{[Y]}  \\ 
&The birth of \textbf{[X]}  occurred in \textbf{[Y]}  \\ 
\midrule[0.4pt]
% \end{tabular}
% \multirow{3}{*}{P190}   
% &The city of \textbf{[X]}   is twinned with \textbf{[Y]}  \\ 
% &\textbf{[X]}   is a twin city of \textbf{[Y]}  \\ 
% &\textbf{[X]}   shares a twinning relationship with \textbf{[Y]}  \\ 
% \midrule[0.4pt]
% \end{tabular}
%  \begin{tabular}{p{0.7cm}|p{5.8cm}}
% \multirow{3}{*}{P20}   
% &\textbf{[X]}  died in \textbf{[Y]}  \\ 
% &\textbf{[X]} 's life ended in \textbf{[Y]}  \\ 
% &Akihiko breathed his last in \textbf{[Y]}  \\ 
% \midrule[0.4pt]
% \multirow{3}{*}{P264}   
% &\textbf{[X]}  is represented by music label \textbf{[Y]}  \\ 
% &In the music world, \textbf{[X]}  is under the label of \textbf{[Y]}  \\ 
% &\textbf{[X]} 's music label is \textbf{[Y]}  \\ 
% \midrule[0.4pt]
% \multirow{3}{*}{P27}   
% &\textbf{[X]}  is a citizen of \textbf{[Y]}  \\ 
% &\textbf{[X]} , who holds a citizenship of \textbf{[Y]}  \\ 
% &The country of citizenship for Rubens Barrichello is \textbf{[Y]}  \\ 
% \end{tabular}
% \begin{tabular}{p{0.7cm}|p{5.8cm}}
% \midrule[0.4pt]
\multirow{3}{*}{\textbf{P276} }         
&\textbf{[X]}  is located in \textbf{[Y]}  \\ 
&\textbf{[X]}  can be found in \textbf{[Y]}  \\ 
&The location of \textbf{[X]}  is \textbf{[Y]}  \\ 
\midrule[0.4pt]
\multirow{3}{*}{\textbf{P36}  } 
&\textbf{[X]}  is the capital of \textbf{[Y]}  \\ 
&\textbf{[X]}  holds the status of being the capital of \textbf{[Y]}  \\ 
&\textbf{[X]} , the capital city of \textbf{[Y]}  \\ 
\midrule[0.4pt]
\multirow{3}{*}{\textbf{P37}   } 
&The official language of \textbf{[X]}  is \textbf{[Y]}  \\ 
&In terms of official language, \textbf{[X]}  uses \textbf{[Y]}  \\ 
&Under \textbf{[X]}  law, the official language is recognized as \textbf{[Y]}  \\ 
\midrule[0.4pt]
\end{tabular}
\begin{tabular}{p{0.7cm}|p{5.8cm}}
\multirow{3}{*}{\textbf{P39}  }
&\textbf{[X]}  assumed the role of \textbf{[Y]}  \\ 
&\textbf{[X]}  holds the position of \textbf{[Y]}  \\ 
&\textbf{[X]}  served in the capacity of \textbf{[Y]}  \\ 
\midrule[0.4pt]
\multirow{3}{*}{\textbf{P407}   } 
&The language of \textbf{[X]}  is \textbf{[Y]}  \\ 
&The \textbf{[X]}  was  penned in \textbf{[Y]}  \\ 
&\textbf{[X]}  was written with the langeuage of \textbf{[Y]}  \\ 
\midrule[0.4pt]
\end{tabular}
\end{table}

\vspace{0.8cm}
~\\

\begin{table}[t!]
\centering
\scriptsize
\begin{tabular}{p{0.7cm}|p{5.8cm}}
\midrule[0.4pt]
\multirow{3}{*}{\textbf{P413}   } 
&\textbf{[X]}  plays in the position of a \textbf{[Y]}  \\ 
&\textbf{[X]} 's role on the team involves him serving as a \textbf{[Y]}  \\ 
&When on the field, \textbf{[X]}  is positioned as a \textbf{[Y]}  \\
\midrule[0.4pt]
\multirow{3}{*}{\textbf{P449}   } 
&\textbf{[X]}  was originally aired on \textbf{[Y]}  \\ 
&\textbf{[X]} , initially, was broadcasted on \textbf{[Y]}  \\ 
&\textbf{[X]}  was originally presented to audiences on \textbf{[Y]}  \\ 
\midrule[0.4pt]
% \multirow{3}{*}{P463}    
% &\textbf{[X]}  belongs to the organization of \textbf{[Y]}  \\ 
% &The organization to which \textbf{[X]}  is a member is \textbf{[Y]}  \\ 
% &\textbf{[X]}  is affiliated with the organization: \textbf{[Y]}  \\ 
% \midrule[0.4pt]
\multirow{3}{*}{\textbf{P495}  }  
&\textbf{[X]}  was created in the country of \textbf{[Y]}  \\    
&The country where creation of \textbf{[X]}  took place is \textbf{[Y]}  \\ 
&\textbf{[X]}  has its birthplace in the country of \textbf{[Y]}  \\ 
\end{tabular}

\begin{tabular}{p{0.7cm}|p{5.8cm}}
\midrule[0.4pt]
\multirow{3}{*}{\textbf{P26}   } 
&\textbf{[X]}  is married to \textbf{[Y]}  \\ 
&The spouse of \textbf{[X]}  is none other than \textbf{[Y]}  \\ 
&In matrimony, \textbf{[X]}  is bound to \textbf{[Y]}  \\  
\midrule[0.4pt]
\multirow{3}{*}{\textbf{P50}  }  
&The author of \textbf{[X]}  is \textbf{[Y]}  \\  
&\textbf{[X]}  was written by \textbf{[Y]}  \\  
&Credited with the creation of \textbf{[X]}  is \textbf{[Y]}  \\  
\midrule[0.4pt]
\multirow{3}{*}{\textbf{P112}  }  
&\textbf{[X]}  was founded by \textbf{[Y]}  \\  
&The \textbf{[X]}  owes its existence to the person of \textbf{[Y]}  \\ 
&The person behind the inception of the \textbf{[X]}  is \textbf{[Y]}  \\  
\end{tabular}

\begin{tabular}{p{0.7cm}|p{5.8cm}}
\midrule[0.4pt]
\multirow{3}{*}{\textbf{P69}   } 
&\textbf{[X]}  was educated at a university named \textbf{[Y]}  \\ 
&\textbf{[X]}  received his education from the institution known as \textbf{[Y]}  \\  
&The university where \textbf{[X]}  was educated is \textbf{[Y]}  \\  
\midrule[0.4pt]
\multirow{3}{*}{\textbf{P140} }   
&\textbf{[X]}  is affiliated with the religion of \textbf{[Y]}  \\  
&\textbf{[X]}  is a follower of the faith known as the \textbf{[Y]}  \\
&The religion that \textbf{[X]}  adheres to is \textbf{[Y]}  \\  
\end{tabular}

 \begin{tabular}{p{0.7cm}|p{5.8cm}}
\midrule[0.4pt]
\multirow{3}{*}{\textbf{P175}}    
&\textbf{[X]}  was performed by \textbf{[Y]}  \\  
&\textbf{[X]} was presented to audiences by \textbf{[Y]}  \\ 
&In the performance of \textbf{[X]} , the artist is \textbf{[Y]}  \\  
\midrule[0.4pt]
\multirow{3}{*}{\textbf{P641}}   
&\textbf{[X]}  is associated with the sport of \textbf{[Y]}  \\  
&The sport that \textbf{[X]}  is linked to is association \textbf{[Y]}  \\  
&\textbf{[X]}  pertains to the sport known as association \textbf{[Y]}  \\  
% \midrule[0.4pt]
% \multirow{3}{*}{P641}    
% &\textbf{[X]}  was created by \textbf{[Y]}  \\  
% &The character of \textbf{[X]}  was the creation of \textbf{[Y]}  \\  
% &The inception of the character \textbf{[X]}  was the work of \textbf{[Y]}  \\  
\bottomrule[0.8pt]
\end{tabular}
\caption{Templates of KLoB-c, where [X] refers to the head entity of fact triples in Wikidata, and  [Y] refers to the tail entity.}
\label{templates}
\end{table}

\vspace{12cm}
~\\

\end{document}